\documentclass[11pt,a4paper]{article}
\usepackage[hyperref]{hesm}
\usepackage{times}
\usepackage{latexsym}

\usepackage{microtype}
\usepackage{graphicx}
\usepackage{amsmath}
\usepackage{amssymb}
\usepackage{amsfonts}
\usepackage{amstext}
\usepackage{amsthm}
\usepackage{comment}

\usepackage{footnote}
\makesavenoteenv{tabular}
\makesavenoteenv{table}

\aclfinalcopy 

\pdfoutput=1

\newcommand{\squishlist}{
	\begin{list}{$\bullet$}
		{ \setlength{\itemsep}{0pt}
			\setlength{\parsep}{3pt}
			\setlength{\topsep}{3pt}
			\setlength{\partopsep}{0pt}
			\setlength{\leftmargin}{1.5em}
			\setlength{\labelwidth}{1em}
			\setlength{\labelsep}{0.5em} } }
	
	\newcommand{\squishlisttwo}{
		\begin{list}{$\bullet$}
			{ \setlength{\itemsep}{0pt}
				\setlength{\parsep}{0pt}
				\setlength{\topsep}{0pt}
				\setlength{\partopsep}{0pt}
				\setlength{\leftmargin}{2em}
				\setlength{\labelwidth}{1.5em}
				\setlength{\labelsep}{0.5em} } }
		
		\newcommand{\squishend}{
	\end{list}  }

\title{Hierarchical Evidence Set Modeling for Automated Fact Extraction and Verification}

\author{Shyam Subramanian {\normalfont{and}} Kyumin Lee\\
	Worcester Polytechnic Institute \\
	Worcester, Massachusetts, 01609, USA \\
	\texttt{\{ssubramanian2,kmlee\}@wpi.edu} \\  
}
\date{}

\begin{document}
\maketitle
\begin{abstract}
Automated fact extraction and verification is a challenging task that involves \emph{finding relevant evidence sentences} from a reliable corpus to \emph{verify the truthfulness of a claim}. Existing models either (i) concatenate all the evidence sentences, leading to the inclusion of redundant and noisy information; or (ii) process each claim-evidence sentence pair separately and aggregate all of them later, missing the early combination of related sentences for more accurate claim verification. Unlike the prior works, in this paper, we propose Hierarchical Evidence Set Modeling (HESM), a framework to extract evidence sets (each of which may contain multiple evidence sentences), and verify a claim to be supported, refuted or not enough info, by encoding and attending the claim and evidence sets at different levels of hierarchy. Our experimental results show that HESM outperforms 7 state-of-the-art methods for fact extraction and claim verification. Our source code is available at \url{https://github.com/ShyamSubramanian/HESM}.

\end{abstract}

\begin{figure*} [t]
    \centering
    \includegraphics[width=0.85\textwidth]{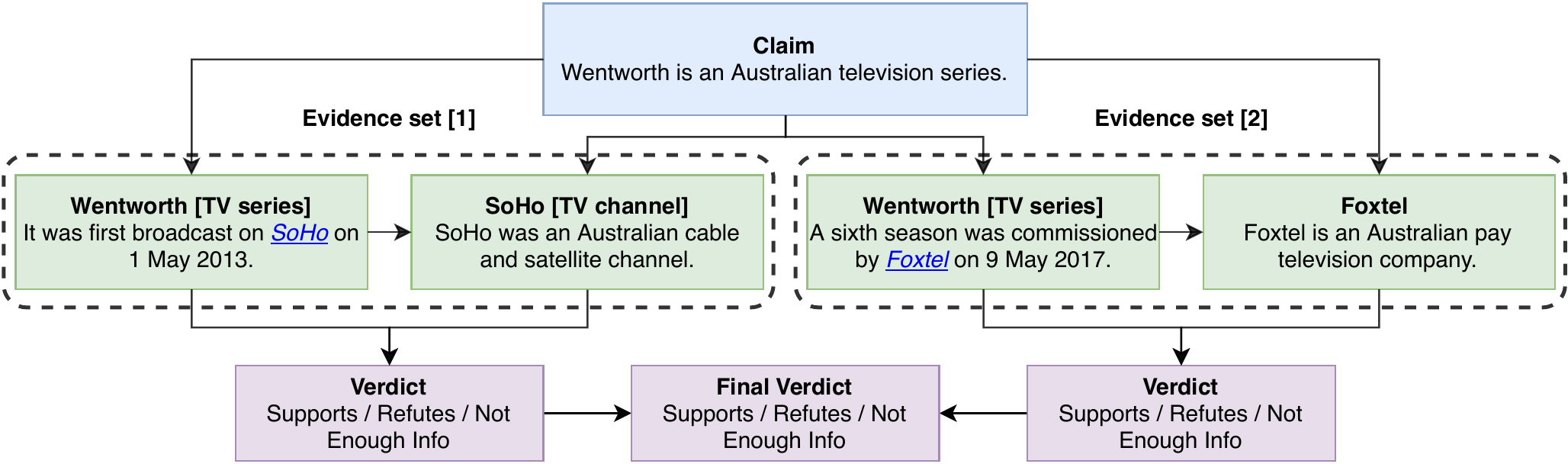}
    \caption{An example of claim, evidence sets, and verdict. The arrows represent the hierarchy of the fact extraction and verification process. The second sentence in each evidence set is retrieved from a document hyperlinked from the first sentence.}
    \label{fig:claim_evidenceset}
\end{figure*}

\section{Introduction}
A study by \citet{gabielkov:hal-01281190} has revealed that 60\% of people on social media share the news after reading just the title, without reading the actual content of the news. Unfortunately, the rise of social media has further accelerated the communication and propagation of unverified information. To solve the problem, our work focuses on automated fact extraction and verification task, which requires retrieving the evidence related to a claim as well as verifying the claim based on the evidence. The task is challenging since it requires semantic understanding and reasoning to learn the subtleties that differ between evidence that supports and evidence that refutes a claim. The task's difficulty is further amplified for claims that require aggregating information from multiple evidence sentences in different documents.

Previous works, in fact verification, either operate by combining all the evidence sentences \cite{Nie_2019} or they operate at each evidence sentence-level and aggregate them later \cite{yoneda-etal-2018-ucl,hanselowski-etal-2018-retrospective}. Combining all the sentences together may lead to the combination of redundant, noisy, and irrelevant information with the relevant information. This makes claim verification more complicated in terms of identifying and learning the context of only the relevant sentences. On the other hand, processing each evidence sentence separately, delays the combination of relevant sentences that belong to the same evidence set, for claims that require aggregating information from multiple sentences. It also makes claim verification harder since it summarizes information without complete context. Figure \ref{fig:claim_evidenceset} depicts an example of an ideal verification system, which extracts evidence sets, processes them individually, and then aggregates them later. In the example, four evidence sentences are retrieved. Sentences which are relevant and hyperlinked, are combined to form evidence sets (called Evidence Set [1] and Evidence Set [2] in the figure). Each evidence set verifies the claim individually, and then they are aggregated for the final verification.

Like Figure \ref{fig:claim_evidenceset}, our proposed framework also retrieves and combines evidence sentences into evidence sets in an iterative fashion. Then, it processes each evidence set individually to form a representation of the evidence set using word-level attention. Then, it combines information from all the evidence set representations using contextual and non-contextual aggregation methods, which use evidence set-level attention. The word-level attention, along with evidence set-level attention, forms a hierarchical attention mechanism. Finally, our framework learns to verify the claim at different levels of hierarchy (i.e., at each evidence set-level and the aggregated evidence level).

Our main contributions are as follows:
\squishlist
\item We propose Hierarchical Evidence Set Modeling, which consists of document retriever, multi-hop evidence retriever, and claim verification.
\item Our multi-hop evidence retriever retrieves evidence sentences and combines them as evidence sets. Our claim verification component conducts the hierarchical verification based on each evidence set individually and then based on all the evidence sets.
\item Our experimental results show that our model outperforms 7 state-of-the-art baselines in both the evidence retrieval and claim verification.
\squishend

\section{Related Work}
Several works exist in fact verification based on different forms of claim and evidence. \citet{thorne-vlachos-2017-extensible,vlachos-riedel-2015-identification} verify numerical claims using subject-predicate-object triples from knowledge graph as evidence. \citet{nakashole-mitchell-2014-language,Bast2017OverviewOT} verify subject-predicate-object triple based claims. \citet{chen2020tabfact} verifies textual claims based on evidences in a tabular format. Fact verification is studied in different natural language settings namely Recognizing Textual Entailment \cite{dagan2015pascal}, Natural Language Inference \cite{snli:emnlp2015} and Claim verification \cite{thorne-etal-2018-fever}. A differently motivated but closely related problem is fact checking in journalism, also known as fake news detection \cite{ferreira-vlachos-2016-emergent,wang-2017-liar}. In this work, we focus on Claim verification using the FEVER dataset \cite{thorne-etal-2018-fever} with textual claims and evidence.

Previous works on the fact extraction and claim verification task follow a three-stage pipeline that includes \textit{document retrieval, evidence sentence retrieval and claim verification}. Most previous works reuse the document retrieval component of top-performing systems \cite{hanselowski-etal-2018-ukp,yoneda-etal-2018-ucl,Nie_2019} in the FEVER Shared Task 1.0 challenge \cite{thorne-etal-2018-fact}.

Evidence sentence retrieval component in almost all previous works retrieves all the evidences through a single iteration \cite{yoneda-etal-2018-ucl,hanselowski-etal-2018-ukp,Nie_2019,chen-etal-2017-enhanced,soleimani2019bert,liu2020fine-grained}. \citet{stammbach-neumann-2019-team} uses a multi-hop retrieval strategy through two iterations to retrieve evidence sentences that are conditioned on the retrieval of other evidence sentences. Then, they choose all the top-most relevant evidence sentences with the highest relevance scores and combine them. Our work follows a similar strategy, but differs from the prior work by combining only evidence sentences that belong to the same evidence set, and then processing each evidence set separately.

In claim verification component, \citet{Nie_2019,yoneda-etal-2018-ucl,hanselowski-etal-2018-ukp} use a modified ESIM model \cite{chen-etal-2017-enhanced} for verification. Recent works \cite{soleimani2019bert,zhou-etal-2019-gear,stammbach-neumann-2019-team} use BERT model \cite{devlin2018bert} for claim verification. Few other works \cite{zhou-etal-2019-gear,liu2020fine-grained} use graph based models for fine-grained semantic reasoning. Different from the previous works, our model operates with claim-evidence set pairs instead of claim-evidence sentence pairs. Our model benefits from encoding, attending and evaluating at different levels of hierarchy, as well as from both contextual and non-contextual aggregations of the evidence sets.

\section{Problem Definition}
Given a set of \textit{\textbf{m}} textual documents and a claim $c_i$, the problem is to find a set of evidence sentences $\hat E_i = \{s_1, s_2, ..., s_{|\hat E_i|}\}$ and classify the claim $c_i$ as $\hat y_i \in \{S,R,NEI\}$ (i.e., SUPPORTED, REFUTED or NOT ENOUGH INFO). For a successful verification of the claim $c_i$, there are two conditions: (1) $\hat E_i$ should match at least one evidence set in the ground truth evidence sets $E_i$ and (2) $\hat y_i$ should match the ground truth entailment label $y_i$.

\section{Hierarchical Evidence Set Modeling}
\begin{figure} [t]
    \centering
    \includegraphics[width=7.5cm]{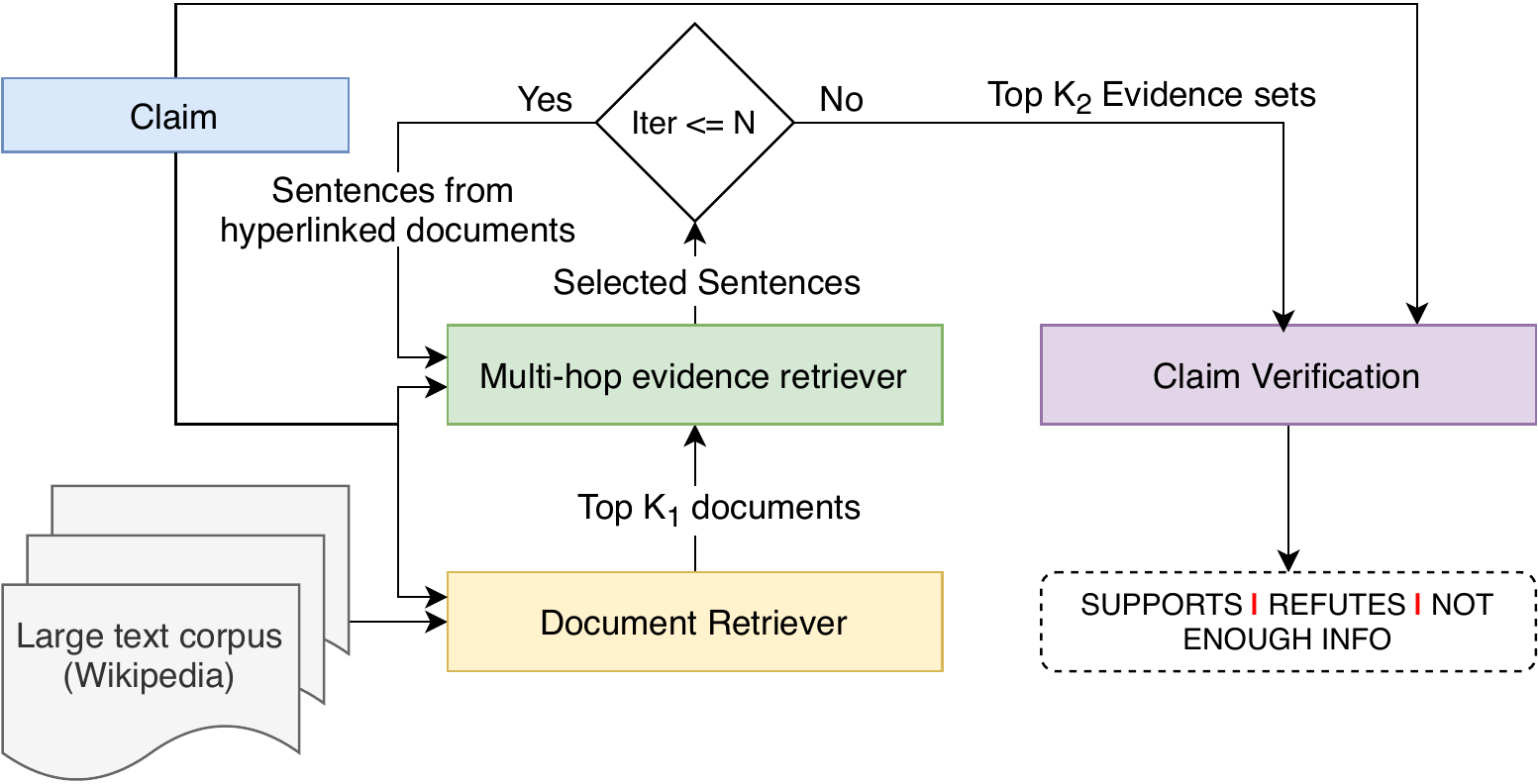}
    \caption{Our HESM framework.}
    \label{fig:hesm_pipeline}
\end{figure}

Our Hierarchical Evidence Set Modeling (HESM) framework consists of three components namely Document Retriever, Multi-hop Evidence Retriever and Claim Verification. Figure~\ref{fig:hesm_pipeline} shows an overview of our framework. The document retriever component retrieves the top $K_1$ documents that are relevant to the claim. The multi-hop retriever component retrieves the relevant top $K_2$ evidence sets from the $K_1$ retrieved documents via an iterative fashion. The claim verification component classifies the claim as SUPPORTS, REFUTES or NOT ENOUGH INFO based on the retrieved evidence sets. Following prior works, in our framework, we reuse the document retriever component from \citet{Nie_2019}, which works well in terms of relevant document retrieval. We mainly focus on and propose novel multi-hop evidence retriever and claim verification components.

\subsection{Document Retriever}
\label{sec:doc_retr}
Document retrieval is the task of selecting documents related to a given claim. First, documents are selected by an exact match between their titles and a span of text of the claim. In particular, the CoreNLP toolkit \cite{manning-EtAl:2014:P14-5} is used for retrieving text spans from the claim. To obtain more relevant documents, the same procedure is applied again after eliminating articles such as 'a', 'an' or 'the' from the claim, and once again after singularizing each word in the claim. For documents, whose titles are ambiguous (e.g., \textit{"Savages (band)"} and \textit{Savages (2012 film)}), a semantic understanding strategy based on \textit{Neural Semantic Matching Network} (NSMN) \cite{Nie_2019} is performed to calculate the relevance of each of the documents by comparing the first line of each document with the claim. Finally, only the top $K_1$ ranked documents are selected.

\subsection{Multi-hop Evidence Retriever}
\label{sec:multi_hop_retr}
According to statistics of the FEVER dataset \cite{thorne-etal-2018-fever}, 16.82\% claims require multiple evidence sentences to verify their truthfulness, and 12.5\% claims' evidence sentences are located across multiple documents. Based on this, we propose a multi-hop evidence retriever, which is an iterative retrieval mechanism with $N$ number of iterations or hops. From analyzing the FEVER dataset, almost all the evidence sentences are at most two hops away from a claim, and thus can be retrieved in two iterations. Hence, for this work, we set $N$ as 2. We retrieve a maximum of $K_2$ evidence sets for each claim. Each evidence set contains a maximum of $M_s$ evidence sentences. With the recent success of Transformer \cite{NIPS2017_7181} based pre-trained models in NLP, we incorporate the ALBERT model \cite{Lan2020ALBERT:} as a part of our multi-hop evidence retriever. ALBERT is a lightweight BERT based model that is pre-trained on large-scale English language corpus for learning language representation.

In the first iteration, given a claim $c_i$, each sentence $j$ in the selected documents from the document retriever is concatenated with the claim $c_i$ as [$[CLS]$;$c_i$;$[SEP]$;$j$] and passed through the ALBERT model. $[CLS]$ and $[SEP]$ are classification and separator tokens required by the ALBERT model. From the ALBERT model representation of each input token, the representation of the $[CLS]$ token is pooled and fed to a linear layer classifier to produce the two scores $m^+$ and $m^-$ for selecting and discarding the sentence, respectively. In Transformer-based models, $[CLS]$ token is considered as a representation of the whole input. Then, a selection probability $p(x=1|c_i,j)$ is calculated as a softmax normalization between the two scores. Only the top $K_2$ sentences with the highest $m^+$ scores and probability score greater than a threshold $th_{evi1}$ are selected.

In the second iteration, each of the $K_2$ evidence sentences from the first iteration is considered as an evidence set. In the FEVER dataset, for claims requiring multiple sentences for verification, most of the sentences missed in the first iteration of retrieval are found in hyperlinked documents of the sentences retrieved in the first iteration. Therefore, in second iteration, the claim $c_i$, each of the $K_2$ evidence sentences $j$, and each sentence $k$ from the hyperlinked documents in sentence $j$ are concatenated as [$[CLS];c_i;[SEP];j;k$] and fed as input to the ALBERT model. Similar to the first iteration, two scores $m^+$ and $m^-$, and a selection probability $p(x=1|c_i,j,k)$ are obtained. Finally, for each evidence sentence $j$, a maximum of ($M_s-1$) sentences with the highest $m^+$ scores and probability score greater than a threshold $th_{evi2}$ are selected and added to the corresponding evidence set.

\subsection{Claim Verification}
\label{sec:hier_evid_set}
Claim verification is a three-way classification task to label the claim as SUPPORTED, REFUTED, or NOT ENOUGH INFO, based on the extracted evidence. Inspired by Hierarchical Attention Network \cite{yang-etal-2016-hierarchical}, we propose a neural network that combines evidence sets hierarchically. While \citet{yang-etal-2016-hierarchical} uses word-level and sentence-level attention to hierarchically combine words into sentences and sentences into a document, in this task, we use word-level and evidence set-level attention to hierarchically combine words and sentences into evidence sets, and evidence sets into an aggregated evidence. Different from \citet{yang-etal-2016-hierarchical}, we propose two ways of aggregating evidence sets. Also, we train each evidence set to be able to verify the claim individually. The model consists of two parts: (1) Evidence Set Modeling Block that contains a word-level encoder and attention layers to model each evidence set based on its words and sentences; and (2) Hierarchical Aggregator that contains evidence set-level encoder and attention layers to combine multiple evidence sets.

\begin{figure} [t]
    \centering
    \includegraphics[width=7.0cm]{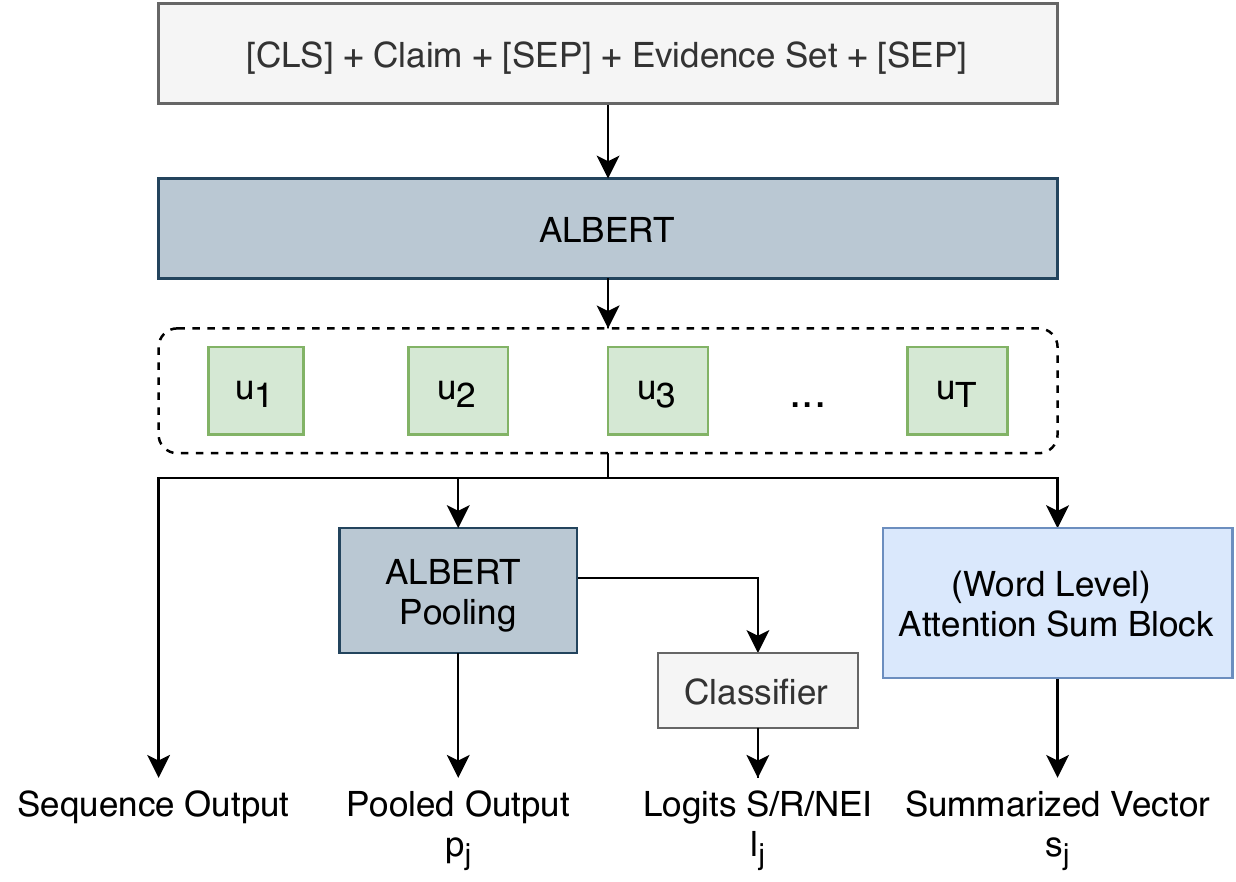}
    \caption{Evidence Set Modeling Block.}
    \label{fig:evid_set_model}
\end{figure}

\subsubsection{Evidence Set Modeling Block}
The Evidence Set Modeling Block in Figure \ref{fig:evid_set_model} takes a claim $c_i$ and each evidence set $e_j$ as input and returns: (1) a sequence output $u_1, u_2, ..., u_T$, that is the representation of each token in the sequence; (2) a pooled output $p_j$, that can be considered as a joint representation of the claim and the evidence set (3) a summarized vector $s_j$, that is also a joint representation of the claim and the evidence set obtained using word-level attention; and (4) the logits $l_j$ from classification of the claim as SUPPORTS, REFUTES or NOT ENOUGH INFO, based on the evidence set $e_j$. 

\textbf{Word Encoder.} We use the ALBERT model for word level encoder. Let $J$ be the number of evidence sets retrieved for the claim $c_i$. First, all the sentences in an evidence set $j$ are concatenated to form the evidence set sequence $e_j$, where $j \in [1,J]$. Then, the claim $c_i$ and the evidence set sequence $e_j$ are concatenated as [$[CLS];c_i;[SEP];e_j;[SEP]$] to form the input sequence $x_j$. The word embeddings, $X_j \in \mathbb{R}^{T \times d}$, of the input sequence $x_{j}$ is obtained from the ALBERT embedding layer, where $T$ denotes the number of tokens in the input sequence $x_{j}$ and $d$ is the size of the word embedding. Then, the ALBERT model processes the input $X_j$ and produces a sequence output $u_1, u_2, ..., u_T$ denoted by $U_j \in \mathbb{R}^{T \times d}$, which consists of the representation of each token $t$ in $x_j$. The ALBERT model also consists of a pooling layer that returns the vector representation $p_j$ of the $[CLS]$ token which is considered to be representation of the whole sequence in Transformer-based models.

\begin{align}
U_j &= \text{ALBERT}(X_j) \in \mathbb{R}^{T \times d} \\
p_j &= \text{ALBERT\_POOLER}(U_j) \in \mathbb{R}^{d}
\end{align}

\begin{figure} [t]
    \centering
    \includegraphics[width=6.5cm,height=6.5cm]{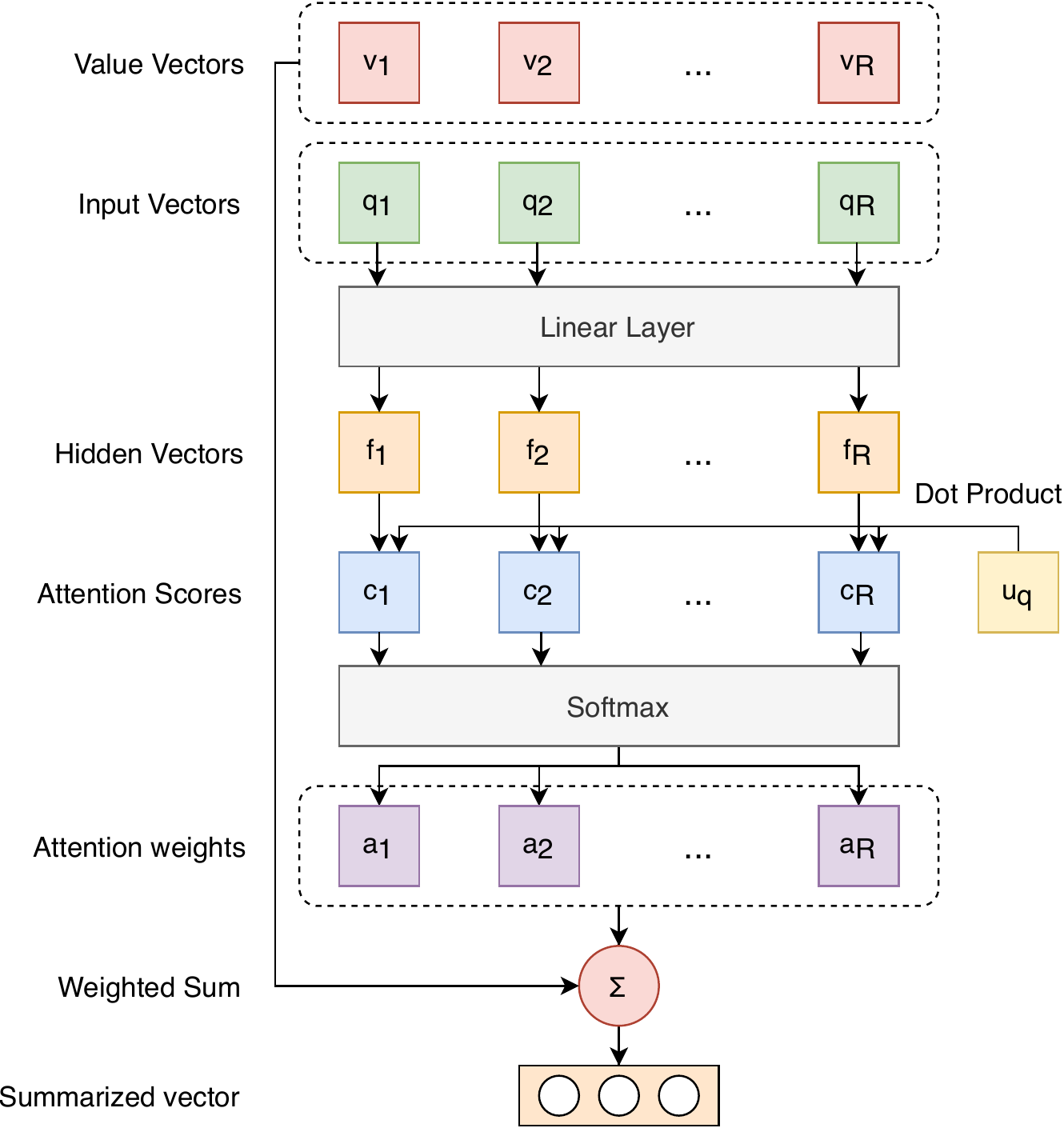}
    \caption{Attention Sum Block.}
    \label{fig:attn_sum}
\end{figure}

\textbf{Attention Sum Block.}
Before describing word-level attention, we first describe the Attention Sum block which is used in the word-level attention. The Attention Sum block in Figure \ref{fig:attn_sum} returns a weighted sum of all the value token vectors $v_1$ to $v_R$, where the weights are calculated using attention between input token vectors $q_1$ to $q_R$ and a trainable weight vector $u_q$ that is randomly initialized. Each vector $q_r$ is passed through a linear layer to get hidden representation $f_r$ for each token $r \in [1,R]$. The hidden representation $f_r$ is then subjected to a dot product with the vector $u_q$ to form a scalar $c_t$ which is the attention score for each $q_r$. Then, softmax is computed over all the attention scores $c_1$ to $c_R$ to get an attention weight $a_r$ for each token $r$. Finally, the value token vectors $v_r$ are subjected to a weighted sum with attention probabilities from the softmax operation as weights and returns the summarized vector $s$. The attention weights denote the importance of each token in the value vectors sequence. The Attention Sum block is used in the following \textit{Word Attention} and \textit{Hierarchical Aggregation} components.

\begin{align}
f_r &= W_qq_r+b_q, r \in [1,R] \\
c_r &= f_r^T {u_q}\\
a_r &= softmax(c_r) \\
s &= \sum_r{v_r {a_r}}
\end{align}

\textbf{Word Attention.} In the word-level attention component, the sequence output $u_t$, where $t \in [1, T]$, of the evidence set $j$ obtained from \textit{Word Encoder} is passed (as both the input $q_r$ and value $v_r$ vectors) through the \textit{Attention Sum block} to obtain a summarized vector representation $s_j$ (denoted as $s$ in Attention Sum block), based on the importance of each word. $s_j$ is used in the \textit{Hierarchical Aggregation} component in Section~\ref{sec:hier_aggr}.

\begin{align}
s_j &= \textnormal{ATTN\_SUM}(u_1, u_2, ..., u_T) \in \mathbb{R}^{d}
\end{align}

\textbf{Classifier.} The pooled output vector $p_j$ containing representation of $[CLS]$ token from the Word Encoder is passed through a linear layer to obtain a three way classification score $l_j$ (SUPPORTS, REFUTES and NOT ENOUGH INFO classes) of the claim $c_i$ based on the evidence set $e_j$. This classifier verifies the claim based on the evidence set.

\begin{align}
    l_j &= W_w{p_j}+b_w
\end{align}

\begin{figure} [t]
    \centering
    \includegraphics[width=7.5cm]{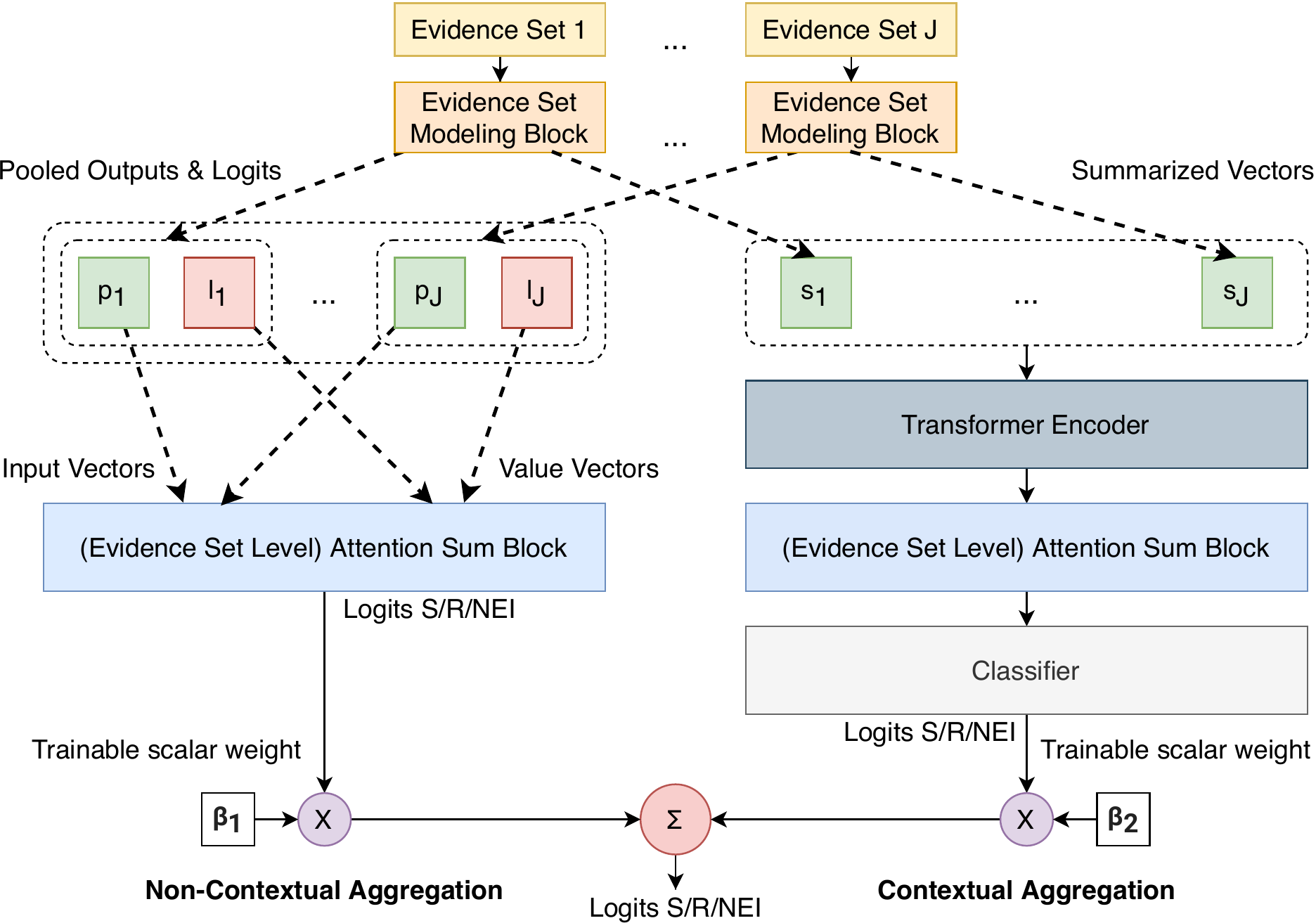}
    \caption{Hierarchical Aggregation.}
    \label{fig:hierarchical_model}
\end{figure}

\subsubsection{Hierarchical Aggregation Modeling}
\label{sec:hier_aggr}
The hierarchical aggregation component in Figure \ref{fig:hierarchical_model} takes the output of the Evidence Set Modeling block of all $J$ evidence sets as input and produces the three-way classification score for the claim based on all the evidence sets. It consists of two types of aggregations namely contextual and non-contextual aggregations. Both components compute an evidence set level attention to combine all the evidence sets, forming a hierarchy.

\textbf{Non-contextual Evidence Set Aggregation.} Non-contextual aggregation combines the logits $l_1, ..., l_J$ of all the evidence sets to produce the aggregated verification logits $l_{nc}$. The motivation behind using non-contextual aggregation is that the majority of the claims need only a single evidence sentence/evidence set for verification. Therefore, we aggregate the logits instead of doing a contextual combination of evidence sets. This helps in avoiding the combination of context from multiple evidence sets without being distracted by sentences containing unnecessary information. The pooled output $p_1, ..., p_J$, and the classification logits $l_1, ..., l_J$ of all the evidence sets, from the Evidence set modeling block, are passed through the Attention Sum block to compute the aggregated representation of all the evidence sets. Here, the sequence of vectors $p_1, p_2, .... p_J$ forms the input vectors of Attention Sum block and the logits $l_1, l_2, ..., l_J$ forms the value vectors of the Attention Sum block. Thus, it aggregates the logits of all evidence sets based on the importance of each evidence set.

\begin{align}
l_{nc} &= \textnormal{ATTN\_SUM}(p_1, ..., p_J; l_1, ..., l_J)
\label{eq:nc_agg}
\end{align}

\textbf{Contextual Evidence Set Aggregation.} Contextual aggregation combines the representation $s_j$ of each evidence set $j$ with one another to produce the claim verification logits $l_c$. The motivation behind using contextual aggregation is that, even though we combine evidence sentences into evidence sets through the multi-hop retriever, our extracted evidence sets might not be completely accurate for some claims (i.e., some evidence sentences that belong to the same ground truth evidence set might be distributed across our extracted multiple evidence sets). Therefore, we combine the evidence sets contextually to overcome the limitation. Let $S \in \mathbb{R}^{J \times d}$ denote the summarized representations $s_1, s_2, ..., s_J$ of all the evidence sets $[1, J]$. $S$ is passed through a Transformer encoder, in order to obtain contextual representations $m_1, m_2, ..., m_J$ denoted by $M \in \mathbb{R}^{J \times d}$. Here, the Transformer encoder layer ensures that the context from one evidence set is combined with other evidence sets. Then, the evidence set representations $m_j$, where $j \in [1, J]$, from the encoder are passed (as both the input $q_r$ and value $v_r$ vectors) through the Attention Sum block to obtain an aggregated vector representation $k$ of all the evidence sets. Finally, the vector representation $k$ is fed into a linear layer classifier to obtain the three way classification logits $l_c$ of the claim.

\begin{align}
M &= \textnormal{Transformer\_Encoder}(S) \in \mathbb{R}^{J \times d} \\
k &= \textnormal{ATTN\_SUM}(m_1, m_2, ..., m_J) \\
l_c &= W_sk+b_s
\label{eq:c_agg}
\end{align}

\textbf{Aggregated Logits.} The aggregated logits are computed based on a weighted combination of the scores from contextual and non-contextual aggregations. The weights $\beta_1$ and $\beta_2$ are trainable weights that denote importance of each aggregation.

\begin{align}
l = \beta_1l_c + \beta_2l_{nc}
\end{align}

\subsubsection{Training Loss and Inference}
\label{sec:loss}
The three-way classification logits $l_j$ from the Evidence set modeling block for each evidence set $j$ are subjected to a cross-entropy loss. All the losses from each evidence set $j$ are averaged to get an aggregated loss $L_{esm}$. The aggregated classification logits $l$ from the Hierarchical Aggregation Modeling block are subjected to a cross-entropy loss $L_{ham}$. The final loss is the sum of $L_{esm}$ and $L_{ham}$.

During the inference, the aggregated logits $l$ from the Hierarchical Aggregation Modeling is used as the final three-way classification score of the claim verification. The label with the maximum score is selected as the final classification label.

\section{Experiments}
\subsection{Experiment Setting}
\label{sec:setting}

In this section, we describe the dataset, evaluation metrics, baselines, and implementation details in our experiments.

\textbf{Dataset.}
We evaluate our framework HESM in the FEVER dataset, a large scale fact verification dataset \cite{thorne-etal-2018-fever}. The dataset consists of $185,445$ claims with human-annotated evidence sentences from $5,416,537$ documents. Each claim is labeled as SUPPORTS, REFUTES, or NOT ENOUGH INFO. The dataset consists of training, development, and test sets, as shown in Table \ref{tab:dataset}. The training and development sets, along with their ground truth evidence and labels are available publicly. But, the ground truth evidence and labels of the test set are not publicly available. Instead, once extracted evidence sets/sentences and predicted labels of the test set by a model are submitted to the online evaluation system\footnote{https://competitions.codalab.org/competitions/18814}, its performance is measured and displayed at the system. In this work, we train and tune our hyper-parameters on training and development sets, respectively. 

\begin{table}[t]
    \centering
    \scalebox{0.7}{
    \begin{tabular}{l | r r r}
    	\hline
    	\textbf{Split} & \textbf{SUPPORTED} & \textbf{REFUTED} & \textbf{NOT ENOUGH INFO}\\
    	\hline
    	\textbf{Train.} & 80,035 & 29,775 & 35,639\\
    	\textbf{Dev.} & 6,666 & 6,666 & 6,666\\
    	\textbf{Test} & 6,666 & 6,666 & 6,666\\
    	\hline
    \end{tabular}}
    \caption{Statistics of FEVER Dataset.}
    \label{tab:dataset}
\end{table}

\textbf{Baselines.}
We compare our model against 7 state-of-the-art baselines, including the top performed models from FEVER Shared task 1.0 \cite{Nie_2019,hanselowski-etal-2018-retrospective,yoneda-etal-2018-ucl}, BERT based models \cite{soleimani2019bert,stammbach-neumann-2019-team,zhou-etal-2019-gear} and graph based model \cite{liu2020fine-grained}. Although we compare ours against all of them, the BERT based models are our major baselines since we use ALBERT, which is a lightweight BERT based model. The detailed description of the baselines is presented in the Appendix.

\textbf{Evaluation Metrics.}
The official evaluation metrics of the FEVER dataset are Label Accuracy (LA) and FEVER score. Label Accuracy is the three-way classification accuracy for the labels SUPPORTS, REFUTES, and NOT ENOUGH INFO, regardless of the retrieved evidence. FEVER score considers a claim to be correctly classified only if the retrieved evidence set matches at least one of the ground truth evidence sets along with the correct label. Between the two metrics, the FEVER score is considered as the most important evaluation metric because it considers both correct evidence retrieval and correct label prediction.

For evidence retrieval performance evaluation, recall and OFEVER are reported since these two scores matter for the claim verification process. Note that OFEVER is the oracle fever score calculated, assuming that the claim verification component has 100\% accuracy. As formulated by \citet{thorne-etal-2018-fever}, a maximum of 5 evidence sentences are extracted to calculate evidence retrieval performance. For our model's evaluation purpose, we assign the score of evidence sentences retrieved in first iteration to their corresponding evidence sets. Then, we sort the evidence sets based on their assigned scores and select at most 5 sentences from the evidence sets in the same sorted order.

\textbf{Implementation, Training, and Hyperparameter Tuning.}
We set number of retrieved documents $K_1=10$, the number of iterations $N=2$, maximum number of sentences retrieved in the first iteration per claim $K_2=3$, a threshold probability $th_{evi1}=0.5$, the maximum number of sentences in each Evidence set $M_{s}=3$, another threshold probability $th_{evi2}=0.8$. Other detailed information is described in the Appendix.

\subsection{Experimental Results and Analysis}
Experiments are conducted to evaluate the performance of evidence retrieval, claim verification, and aggregation approaches. In addition, we conduct an ablation study. Only the claim verification experiment is conducted in the test set since each baseline's officially evaluated results are reported in the FEVER leaderboard. In the other experiments and analysis, we use the development set since the test set does not contain the ground truth of evidence sets/sentences and claim class labels.

\begin{table}[t]
    \centering
    \scalebox{0.53}{%
    \begin{tabular}{l | c | c c}
    	\hline
    	Model & \# of Iterations & Recall & OFEVER (\%)\\
    	\hline
    	UNC NLP \citet{Nie_2019} & 1 & 0.868 & 91.19\\
    	BERT-Base \citet{stammbach-neumann-2019-team} & 2 & 0.898 & 93.20\\
    	\emph{our} HESM (ALBERT-Base) & \textbf{2} & \textbf{0.905} & \textbf{93.70}\\
    	\hline
    \end{tabular}}
    \caption{Evidence retrieval performance of the baselines and our model in development set.}
    \label{tab:mhe}
\end{table}

\subsubsection{Multi-hop evidence retrieval}
As shown in Table \ref{tab:mhe}, we compare the performance of our model with two baselines, UNC NLP \cite{Nie_2019} and BERT based model \cite{stammbach-neumann-2019-team}. UNC NLP uses ESIM \cite{chen-etal-2017-enhanced} based model, and \citet{stammbach-neumann-2019-team} uses a BERT based model. Since most other previous works either use ESIM based model or BERT based model for evidence retrieval, we compare with these two representative baselines (i.e., the results of the other 5 baselines in evidence retrieval would be similar to one of them). Our HESM with ALBERT Base outperforms the baselines, achieving 0.905 recall and 93.70\% OFEVER score. We can also notice that multiple-hop evidence retrieval approaches (ours and \citet{stammbach-neumann-2019-team}) performed better than UNC NLP, which conducts a single iteration.

\begin{table}[t]
    \centering
    \scalebox{0.65}{%
    \begin{tabular}{l | c c c}
    	\hline
    	Model & LA(\%)& FEVER(\%)\\
    	\hline
    	UKP Athene \cite{hanselowski-etal-2018-ukp} & 65.46 & 61.58\\
    	UCL MRG \cite{yoneda-etal-2018-ucl} & 67.62 & 62.52\\
    	UNC NLP \cite{Nie_2019} & 68.21 & 64.21\\
    	\hline
    	BERT Pair \cite{zhou-etal-2019-gear} & 69.75 & 65.18\\
    	BERT Concat \cite{zhou-etal-2019-gear} & 71.01 & 65.64\\
    	BERT (Base) \cite{soleimani2019bert} & 70.67 & 68.50\\
    	GEAR (BERT Base) \cite{zhou-etal-2019-gear} & 71.60 & 67.10\\
    	KGAT (BERT Base) \cite{liu2020fine-grained} & 72.81 & 69.40\\
    	\emph{our} HESM (BERT Base) & \textbf{73.18} & \textbf{70.07}\\
    	\emph{our} HESM (ALBERT Base) & \textbf{73.25} & \textbf{70.06}\\
    	\hline
    	BERT (Large) \cite{soleimani2019bert} & 71.86 & 69.66\\
    	BERT (Large) \cite{stammbach-neumann-2019-team} & 72.71 & 69.99\\
    	KGAT (BERT Large) \cite{liu2020fine-grained} & 73.61 & 70.24\\
    	KGAT (RoBERTa Large) \cite{liu2020fine-grained} & 74.07 & 70.38\\
    	\emph{our} HESM (ALBERT Large) & \textbf{74.64} & \textbf{71.48}\\
    	\hline
    	\end{tabular}}
    \caption{Performance of the baselines and our model in test set.}
    \label{tab:ver_test}
\end{table}

\subsubsection{Claim verification}
Table \ref{tab:ver_test} shows claim verification results of our HESM model and baselines. Our model with ALBERT Large outperforms all the baselines, achieving 74.64\% label accuracy (LA) and 71.48\% FEVER score. In particular, our model performed much better than the top performed models from FEVER Shared task 1.0 (i.e., UKP Athene, UCL MRG, and UNC NLP). Compared with baselines using BERT Base, our HESM with ALBERT Base performed better than them. Likewise, compared with baselines using large language models, our model with ALBERT large still performed better than them. This experimental result confirms that our model with ALBERT large improved 1.1\% FEVER score compared with the best baseline, KGAT with RoBERTa Large, indicating our model's capability of producing more correct label prediction and evidence extraction. The reason why we chose to use ALBERT over BERT in our models is ALBERT consumes much less memory and is expected to have comparable performance to its BERT counterpart. Since the other models/baselines use BERT instead of ALBERT, for a fair comparison, we include a result of our HESM model with BERT Base. The performance is similar to the HESM with ALBERT Base model. This result confirms that our framework is more important than a specific language model used.

\begin{table}[t]
    \centering
    \scalebox{0.75}{%
    \begin{tabular}{l | c c}
    	\hline
    	Aggregation & LA(\%) & FEVER(\%)\\
    	\hline
    	Logical & 68.92 & 66.32\\
    	Top-1 & 69.92 & 67.77\\
    	MLP & 74.25 & 72.03\\
    	Concat & 74.87 & 72.13\\
    	Attention-based & 74.96 & 72.74\\
    	HESM & \textbf{75.77} & \textbf{73.44}\\
    	\hline
    \end{tabular}}
    \caption{Claim verification with different aggregation methods in development set.}
    \label{tab:aggr}
\end{table}

\subsubsection{Aggregation Analysis}
We compare our hierarchical aggregation with different baseline aggregation methods. Table \ref{tab:aggr} shows the results of aggregation analysis in the development set. Top-1 aggregation is using just the top-1 relevant evidence set to verify the claim. Logical aggregation involves classifying the claim as SUPPORTS or REFUTES if at least one of the evidence sets has the label SUPPORTS or REFUTES, respectively. In case both labels appear in the evidence sets, then the label from the top-scoring evidence set is used to break the tie. If both labels do not appear in any of the evidence sets, we predict the claim as NOT ENOUGH INFO. MLP aggregation is to use an MLP layer to aggregate the class label probability of all the evidence sets to get a final verification label. Concat aggregation concatenates all the sentences in all evidence sets into a string to verify the claim. Attention-based aggregation is similar to the aggregation technique used in \citet{hanselowski-etal-2018-ukp} using attention between claim and each evidence set to get the importance of each evidence set and then combine them using Max and Mean pooling. Finally, our HESM model aggregates evidence sets using hierarchical aggregation. From the results, we can see that our HESM model outperforms all other aggregation methods.

\begin{table}[t]
    \centering
    \scalebox{0.8}{
    \begin{tabular}{l | c c}
    	\hline
    	Model & LA(\%) & FEVER(\%)\\
    	\hline
    	HESM & 75.77 & 73.44\\
    	- w/o Evidence set level Loss & 75.35 & 72.74\\
    	- w/o Non-Contextual Aggregation & 75.33 & 72.74\\
    	- w/o Contextual Aggregation & 73.70 & 71.96\\
    	- w/o Multi-hop evidence retrieval & 73.53 & 71.92\\
    	\hline
    \end{tabular}}
    \caption{Ablation analysis in development set. }
    \label{tab:ablation}
\end{table}

\subsubsection{Ablation Study}
Table \ref{tab:ablation} shows the label accuracy and FEVER score of our model after removing different components, including evidence set level loss $L_{esm}$, and contextual and non-contextual aggregations. All of the proposed components positively contributed to boosting the performance of our framework.

\begin{figure}[t]
    \centering
    \includegraphics[width=7.5cm]{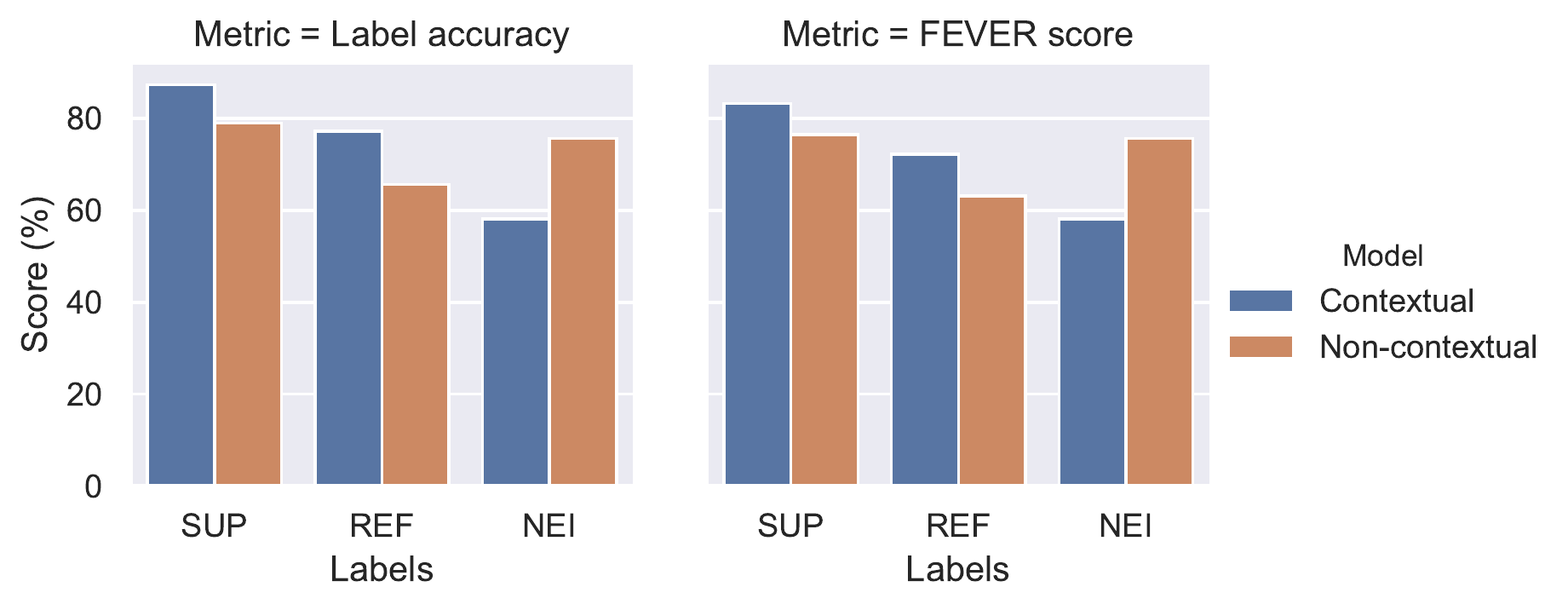}
    \caption{Performance of contextual and non-contextual aggregations given different claim labels.}
    \label{fig:c_nc_labels}
\end{figure}

\begin{figure} [t]
    \centering
    \includegraphics[width=7.5cm]{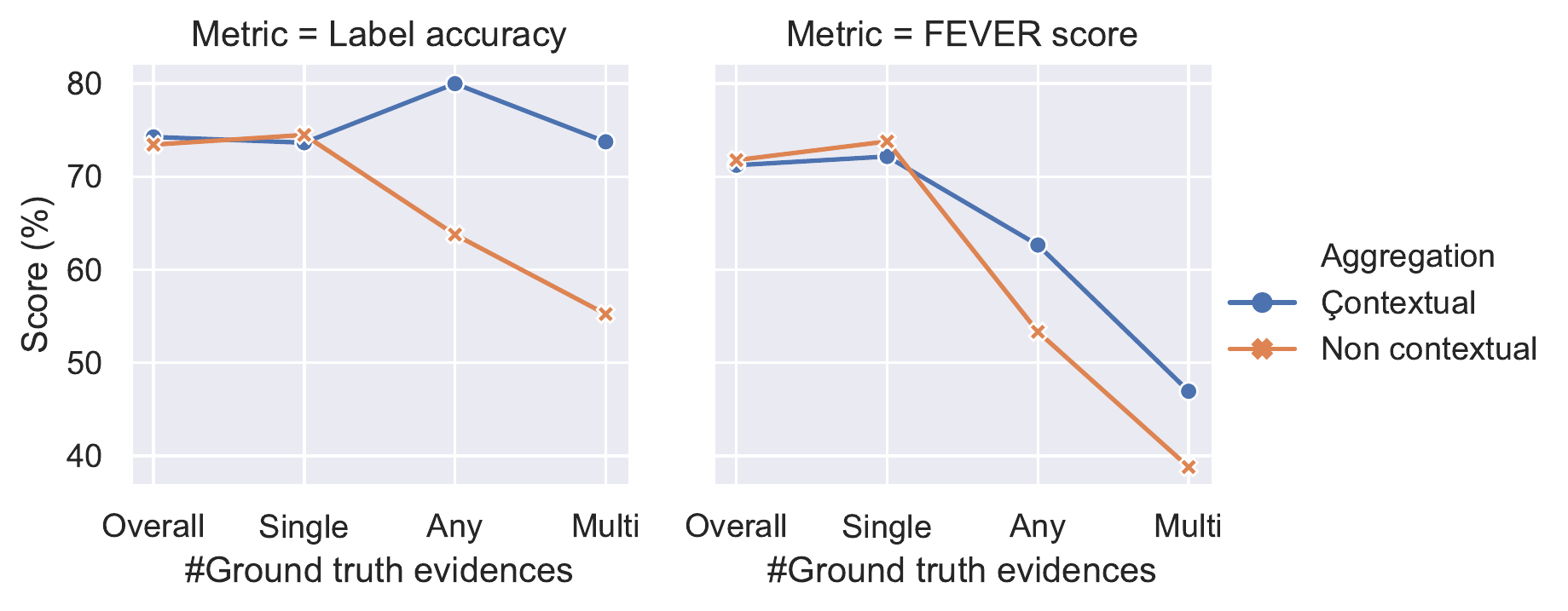}
    \caption{Performance of contextual and non-contextual aggregations given claims requiring different number of evidence sentences.}
    \label{fig:c_nc_gtevid}
\end{figure}

\subsubsection{Contextual and Non-contextual Aggregations}
\label{sec:nc_v_c}
In this section, we study the performance of contextual and non-contextual aggregations in different aspects in the development set.

\textbf{Label-wise performance.} Figure \ref{fig:c_nc_labels} shows performance of contextual and non-contextual aggregations with respect to the class labels. We use the logits $l_c$ and $l_{nc}$ to calculate performance of contextual and non-contextual aggregations. In both label accuracy and FEVER score, contextual aggregation performs better for correctly verifying a claim when the relevant evidence either supports or refutes the claim, whereas non-contextual aggregation performs better in identifying evidence that does not have enough information to support or refute the claim (i.e., claims with the label NOT ENOUGH INFO). Thus, each aggregation complements the other in claim verification.

\textbf{Performance on claims requiring a different number of evidence sentences.} Figure \ref{fig:c_nc_gtevid} shows the performance of contextual and non-contextual aggregations with respect to claims requiring a different number of evidence sentences for verification. \textit{Overall} refers to all the claims, \textit{Single} refers to claims requiring only a single evidence sentence for verification, \textit{Any} refers to claims for which more than one ground truth evidence set exists, where some sets contain a single evidence sentence and some sets contain multiple evidence sentences, and \textit{Multi} refers to claims that can be verified only with multiple sentences. Non-contextual aggregation performs better than contextual aggregation in claims requiring only \textit{Single} evidence sentence, whereas contextual aggregation performs better than non-contextual aggregation in claims requiring \textit{Any} and \textit{Multi} evidence sentences. The results make sense because contextual aggregation combines the context of multiple evidence sets, while non-contextual aggregation usually selects one of the evidence sets based on the attention mechanism.

\begin{table}[t]
    \centering
    \scalebox{0.8}{%
    \begin{tabular}{l | c c}
    	\hline
    	Aggregation & Weights & Attention accuracy \\
    	\hline
    	Contextual & 0.48 & 90.93\\
    	Non-contextual & 0.52 & 92.41 \\
    	\hline
    \end{tabular}}
    \caption{Attention analysis for contextual and non-contextual aggregation}
    \label{tab:c_nc_metrics}
\end{table}

\textbf{Attention analysis.} In Table \ref{tab:c_nc_metrics} we show the weights $\beta_1$ and $\beta_2$ of the final model and also the evidence-set level attention accuracy. The attention weights can be seen as the importance of each aggregation. The attention weights show that both the aggregations are equally important (0.48 vs. 0.52).

The attention accuracy denotes the accuracy of the evidence set-level attention from the attention sum block in both eq. (\ref{eq:nc_agg}) and eq. (\ref{eq:c_agg}) of non-contextual and contextual aggregations, respectively. It evaluates whether the retrieved evidence set from multi-hop retriever, which matches one of the ground truth evidence sets, has the highest attention weight of all the retrieved evidence sets. In cases where the evidence sentences from the ground truth evidence set are distributed across multiple evidence sets retrieved from multi-hop retriever, the attention is considered accurate if all the matching evidence sets have higher attention weight than the non-matching evidence sets. Here, we consider only the claims for which the retrieved evidence sentences match at least one ground truth evidence set. In other words, we omit the claims with NOT ENOUGH INFO label and also the 6.3\% claims for which the multi-hop retriever cannot retrieve evidence sentences that match at least one ground truth evidence set as shown in Table \ref{tab:mhe}. The high attention accuracy for both contextual and non-contextual aggregation shows that our evidence-set level attention is highly capable of attending to the correct evidence sets.

\section{Conclusion}
In this paper, we have proposed HESM framework for automated fact extraction and verification. HESM operates at evidence set level initially and combines information from all the evidence sets using hierarchical aggregation to verify the claim. Our experiments confirm that our hierarchical evidence set modeling outperforms 7 state-of-the-art baselines, producing more accurate claim verification. Our aggregation and ablation study show that our hierarchical aggregation works better than many baseline aggregation methods. Our analysis of contextual and non-contextual aggregations illustrates that the aggregations perform different roles and positively contribute to different aspects of fact-verification. 

\section*{ACKNOWLEDGEMENTS}
This work was supported in part by NSF grant CNS-1755536, AWS Cloud Credits for Research, and Google Cloud. Any opinions, findings, and conclusions or recommendations expressed in this material are the author(s) and do not necessarily reflect those of the sponsors.

\bibliographystyle{acl_natbib}
\bibliography{hesm}

\clearpage
\appendix
\section{Appendix}

\textbf{Baselines.}
We compare our model with 7 state-of-the-art baselines, including the top performed models from FEVER Shared task 1.0 \cite{thorne-etal-2018-fact}, BERT based models, and a graph-based model.

The top performed models from FEVER shared task 1.0 include UNC NLP \cite{Nie_2019}, UKP Athene \cite{hanselowski-etal-2018-retrospective} and UCL MRG \cite{yoneda-etal-2018-ucl}. All three models use a modified version of Enhanced Sequential Inference Model \cite{chen-etal-2017-enhanced} for claim verification. UNC NLP model concatenates all retrieved evidence sentences together to verify the claim, whereas UCL MRG and UKP Athene models process each evidence sentence separately and aggregate them at a later stage. UCL MRG reports the best results with linear layer aggregation. UKP Athene uses an attention-based aggregation. 

The BERT based models include \citet{soleimani2019bert,stammbach-neumann-2019-team,zhou-etal-2019-gear}. \citet{soleimani2019bert} uses BERT-base and BERT-large for evidence retrieval and claim verification, respectively. They also experiment with both pairwise and point-wise ranking for evidence retrieval. \citet{stammbach-neumann-2019-team} uses two iterations of evidence retrieval similar to our work, but different from our work, they concatenate all the sentences retrieved. \citet{zhou-etal-2019-gear} reports performance for both BERT-concat that concatenates all the sentences and BERT-pair model that processes each claim-evidence sentence pair separately. GEAR \cite{zhou-etal-2019-gear} uses BERT Base as backbone and aggregates claim-evidence sentence pair using a fully-connected graph-based evidence reasoning network. A graph-based model KGAT \cite{liu2020fine-grained} uses a modified version of Graph Attention Network \cite{velikovi2018graph} to model a graph constructed from claim and evidence. KGAT experiments with both BERT Base and BERT Large models as its backbone.

\textbf{Detailed Implementation, Training and Hyperparameter Tuning.}
For training the document retriever, Adam optimizer \cite{kingma2014adam} is used with a batch size of 128, and cross-entropy loss is used. The maximum number of retrieved documents $K_1$ is set to 10. In the Multi-hop evidence retrieval stage, the number of iterations $N$ is set to 2. For both iterations, the ALBERT-Base model for sequence classification is used and is trained using a batch size of 64 along with AdamW optimizer \cite{loshchilov2018decoupled} and a learning rate of 5e-5. In the first iteration, we set the threshold probability $th_{evi1}$ as 0.5, and the maximum number of sentences per claim $K_2$ to 3. We also use the annealed sampling strategy followed by \citet{Nie_2019} to decrease the number of negative examples after each epoch so that model learns to be more tolerant about selecting sentences while being discriminative enough to filter out apparent negative sentences.

In the second iteration, we use the ALBERT-Base model to retrieve relevant sentences in hyperlinked documents of evidence sentences retrieved in the first iteration. Similar to the first iteration, we use annealed sampling here as well. We set the maximum number of sentences in an Evidence set, $M_{s}$ to be 3. Finally, we choose either $K_2$ evidence sets or lesser, if a lesser number of evidence sets leads up to 5 evidence sentences since only 5 evidence sentences are considered for calculating FEVER score.  We set the threshold probability $th_{evi2}$ to 0.8 since we find that the model is able to retrieve correct evidence sentences with a high probability. Cross entropy loss is used in both iterations. Both the iterations are trained for 4 epochs.

Finally, in the claim verification stage, we use the hierarchical evidence set aggregator, which uses the ALBERT model as its backbone. We use AdamW optimizer with a batch size of 32 and a learning rate of 2e-5 and 4 epochs to train our final model. It also uses a 2 layer transformer encoder for evidence-set level encoding. 

We use the PyTorch framework to optimize both Multi-hop evidence retriever and claim verification components. We use grid-search on development set to search over a batch size from \{32, 64\}, a learning rate from \{2e-5, 5e-5\}, and number of epochs from \{2, 4, 6\}. The maximum number of evidence sets $K_2$ is selected from \{2, 3, 4\} and maximum number of sentences per evidence set $M_s$ is selected from \{2, 3, 4\}. In claim verification, the number of transformer encoder layers in contextual aggregation is selected from \{1, 2, 3\}.
\end{document}